\pdfoutput=1

\documentclass[11pt]{article}

\usepackage[preprint]{acl}

\usepackage{times}
\usepackage{latexsym}

\usepackage[T1]{fontenc}

\usepackage[utf8]{inputenc}

\usepackage{microtype}

\usepackage{graphicx}
\usepackage{subfigure}
\usepackage{booktabs}
\usepackage{multirow}
\usepackage{multicol}
\usepackage{wrapfig}
\usepackage{float}
\usepackage{colortbl}
\usepackage{xcolor}
\usepackage{inconsolata}
\usepackage{arydshln}
\usepackage{makecell}
\usepackage[misc,geometry]{ifsym}

\usepackage{dsfont}
\usepackage{amsmath}
\definecolor{myblue}{HTML}{5BCEFA}
\definecolor{mypink}{HTML}{F5A9B8}
\definecolor{myyellow}{HTML}{FFB47B}
\definecolor{ppurple}{HTML}{63b931}
\usepackage{pgfplots}
\usetikzlibrary{pgfplots.groupplots}
\pgfplotsset{compat=1.3}
\usepackage{tikz}
\usepackage[capitalize,noabbrev]{cleveref}
\usetikzlibrary{patterns}

\usepackage{hyperref}

\definecolor{myYellow}{rgb}{0.9,0.9,1}

\DeclareMathOperator*{\myargmax}{argmax} 

\newcommand{\prettytext}{\textsc{Pretty}}

\newcommand\blfootnote[1]{%
  \begingroup
  \renewcommand\thefootnote{}\footnote{#1}%
  \addtocounter{footnote}{-1}%
  \endgroup
}

\title{Prefix Text as a Yarn:\\Eliciting Non-English Alignment in Foundation Language Model}
\author{Runzhe Zhan$^{\clubsuit\text{*}}$~~
        Xinyi Yang$^{\clubsuit}$~~
        Derek F. Wong$^{\clubsuit\text{~\Letter}}$~~
                \textbf{Lidia S. Chao}$^{\clubsuit}$~~Yue Zhang$^{\spadesuit\text{~\Letter}}$\\
  $^{\clubsuit}$NLP$^2$CT Lab, Department of Computer and Information Science, 
  University of Macau\\
  $^{\spadesuit}$School of Engineering, Westlake University \\
  \texttt{nlp2ct.\{runzhe, xinyi\}@gmail.com, \{derekfw, lidiasc\}@um.edu.mo} \\
          \texttt{zhangyue@westlake.edu.cn} 
}

\begin{document}
\maketitle
\blfootnote{$~~^\text{*}$Work was done during a visit to Westlake University.}
\blfootnote{$^\text{\Letter}$~Co-corresponding authors.}

\begin{abstract}
While supervised fine-tuning (SFT) has been a straightforward approach for tailoring the output of foundation large language model (LLM) to specific preferences, concerns have been raised about the depth of this alignment, with some critiques suggesting it is merely ``superficial''.
We critically examine this hypothesis within the scope of cross-lingual generation tasks, proposing that the effectiveness of SFT may be constrained by its reliance on prior tokens to guide cross-lingual generation.
Based on this crucial insight, and in response to the challenges posed by the costly and limited availability of non-English data for SFT, we introduce a novel training-free alignment method named \prettytext, which employs minimal task-related prior tokens to bridge the foundation LLM and the SFT LLM, achieving comparable performance without training. Experiments on machine translation and part-of-speech tagging across eight languages demonstrate the efficacy of \prettytext~in cross-lingual settings. Remarkably, by initiating the decoding process with only one or two prior tokens, foundation LLMs can achieve performance comparable to their SFT counterparts. This method presents a cost-effective alternative to SFT and advances the democratization of multilingual LLMs.
\end{abstract}
 
\section{Introduction}
Supervised fine-tuning (SFT) refines large language models (LLMs) using task-specific instruction data to enhance their capability to follow instructions \cite{touvron2023llama, peng2023instruction} and to align their outputs with human preferences and safety considerations \cite{ouyang2022training, rafailov2023direct, dong2023raft, yuan2023rrhf}.
This process is often termed ``alignment'', signifying the tailoring of model outputs to conform to specific downstream requirements. Nevertheless, current research casts doubt on the necessity and potential adverse impacts of SFT. 
But the alignment achieved through SFT is often considered to be ``superficial'', with the process potentially repurposing pre-existing knowledge from pre-training to merely reshape outputs to meet specific criteria \cite{zhou2023lima, lin2023unlocking}. 
It has been observed that even a small-scale SFT training dataset can produce significant alignment effects \cite{liu2023makes, xia2024less}.
On the other hand, recent empirical studies \cite{luo2023empirical, dong2023abilities} have raised concerns that SFT might hurt the knowledge acquired during its pre-training phase, leading to serious consequences like catastrophic forgetting. 

Not only is there no definitive consensus on the necessity of SFT, but the majority of these studies also focus on monolingual tasks. LLMs still encounter challenges in handling complex cross-lingual generation tasks \cite{ DBLP:journals/corr/abs-2305-11778, wang-etal-2023-zero}. 
Current research on cross-lingual alignment primarily seeks to extrapolate or align English capabilities to other languages using the SFT paradigm \cite{DBLP:journals/corr/abs-2306-10968, DBLP:journals/corr/abs-2401-07037, DBLP:journals/corr/abs-2403-11621}, yet there remains a gap in exploring the specific impacts of SFT-based cross-lingual alignment. Furthermore, given the potential risk of SFT leading to the forgetting of pre-training knowledge, the question of how to achieve cross-lingual alignment without training remains underexplored.

\begin{figure*}
    \centering
    \includegraphics[scale=0.24]{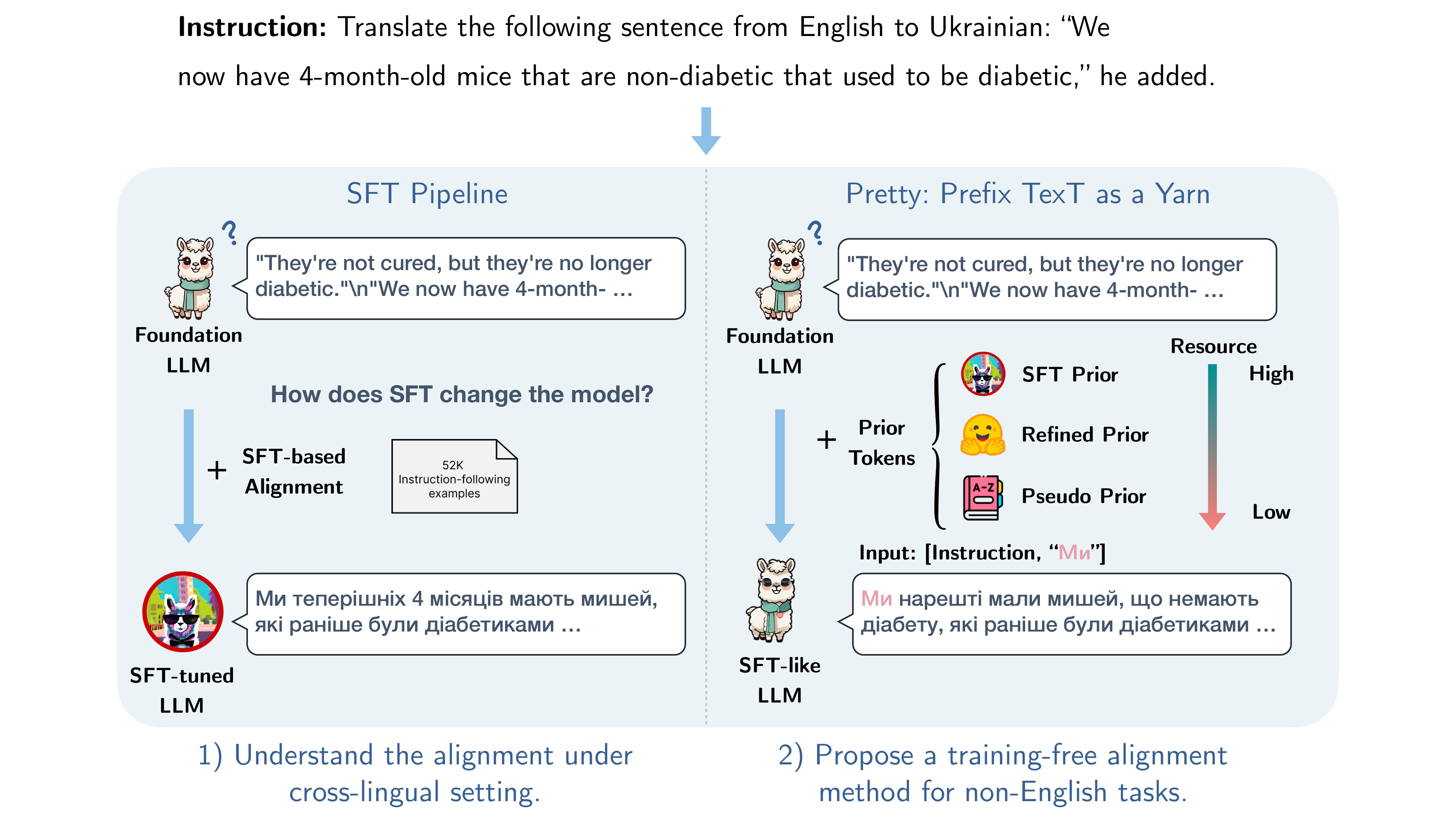}
    \caption{Illustration of our research question and proposed \textbf{P}refix \textbf{T}ex\textbf{T} as a \textbf{Y}arn (\prettytext) framework.}
    \label{fig:main-fig}
\end{figure*}

To bridge these gaps, our study conducts an in-depth examination of the impact of SFT on cross-lingual generation. We investigate the influence of SFT on the decoding patterns of foundation models in cross-lingual contexts, hypothesizing that the success of SFT largely hinges on the selection of initial prior tokens that are critical for eliciting task-specific generation in the target language.
Furthermore, the observed decoding similarities between foundation and SFT models support the extension of the superficial alignment hypothesis to cross-lingual scenarios.
Responding to these insights, we introduce a training-free alignment method named ``\prettytext'' for cross-lingual and non-English tasks. The \textbf{Pre}fix \textbf{T}ex\textbf{T}s act as a \textbf{Y}arn (\prettytext) linking the foundation LLM and the SFT LLM, eliciting the foundation LLM to exhibit near-SFT performance levels. 
Specifically, we augment the original input with a few tokens that serve as decoding priors, and then prompt the foundation LLM to resume decoding based on this modified input.
In most cases, only one or two task-related prior tokens are needed, and the method for constructing these prior tokens is flexible across various kinds of language resources, fostering the democratization of multilingual LLMs.

We conducted experiments on machine translation \cite{goyal2022flores}, cross-lingual summarization \cite{bhattacharjee-etal-2023-crosssum} and non-English part-of-speech (POS) tagging \cite{liang-etal-2020-xglue} tasks across eight languages.
These tasks exemplify cross-lingual generation and multilingual language understanding, and they provide ample non-English test data to evaluate effectiveness across varying levels of resource availability.
The experimental results demonstrate that $\prettytext$ can effectively align the foundation model to match SFT model's performance without training, by merely adding two prior tokens in the decoding.
\section{Iceberg Model of SFT}

\subsection{Preliminaries}\label{sec:prelimi}
\paragraph{Pre-training} 
The pre-training (PT) of LLMs is primarily conducted through language modeling tasks on large-scale unlabeled data \cite{touvron2023llama, achiam2023gpt}. 
During this phase, given a sequence $X_\mathrm{PT}$ of length $N$ and a context window $k$, the optimization objective is maximizing the joint probability $P_\mathrm{LM}$ as:
\begin{equation}\label{eq:pt}
    P_\mathrm{LM}(X_\mathrm{PT})=\prod_{i=1}^N P(x_i|x_{i-k:i-1})
\end{equation}
which encourages the model to generate text that naturally follows from the preceding context. However, this ``text completion'' behavior can become a bottleneck when models are prompted to switch languages or follow specific instructions of cross-lingual generation. 
It is frequently observed that when prompted with English input and instructed to produce text in a different language, as illustrated in the upper example of \cref{fig:main-fig}, the foundation model often continues to decode in English.

\paragraph{SFT} 
SFT leverages labeled data pair ${(X_\mathrm{ins.}, Y)}$ to empower models with the ability to follow instructions. This stage aims to maximize the probability of the expected answer $Y$ conditioned on the input text $X_\mathrm{ins.}$, where $X_\mathrm{ins.}$ consists of the task instruction and task input.
\begin{equation}\label{eq:sft}
    P_\mathrm{SFT}(Y|X_\mathrm{ins.})= 
 \prod_{j=1}^T P(y_j|y_{1:j-1},X_\mathrm{ins.})
\end{equation}
SFT is crucial for aligning foundation models to perform task-specific instructions, effectively transforming a general-purpose LLM into an instruction-following assistant.
However, data quality, training costs, and the imbalance of multilingual data hinder the democratization of assistant LLM. As mentioned before, SFT may be harmful to pre-training knowledge. Thus, it is meaningful and important to understand the underlying mechanism of SFT-based alignment and propose a more efficient alignment method.

\subsection{Beneath the SFT-based Alignment}
\paragraph{Prior Knowledge Hypothesis}
It is worth noting that pre-training corpora also contain sequences that naturally express task-specific information, which imparts certain capabilities to the foundation LLMs. 
For example, the presence of semantically equivalent expressions in the pre-training text may enable LLM acquire machine translation ability during pre-training stage \cite{radford2019language}.

Despite its extensive prior knowledge, the foundation LLM still struggles with complex cross-lingual generation tasks. Beyond existing studies, we provide more concrete insights into this issue by prompting foundation LLMs with various instructions \cite{bawden-yvon-2023-investigating}. Notably, only 31.8\% of these prompts successfully elicit translation capability from the foundation LLMs\footnote{For detailed information, please refer to \cref{sec:success}.}.

This deficiency may stem from two main factors: First, the proportion of text with the aforementioned characteristics in the pre-training corpus $X_\mathrm{PT}$ is still relatively small, and most of it is far from resembling human instruction text $X_\mathrm{ins.}$.  
Consequently, the model is more likely to predict tokens suitable for completing formal texts than those required for task-specific instructions. As a result, the foundation LLM often fails to produce tokens $y \in Y_{1:T}$ in the intended target language.
Secondly, the predominance of English in the pre-training data skews the token generation probabilities of foundation LLM. Given a cross-lingual context, the model favors predicting tokens in English, while the token probabilities for other languages remain comparatively low. For example, English data comprises up to 90\% of the Llama2 pre-training data \cite{touvron2023llama}, which may lead models to generate text with an English-centric bias. 

The above hypothesis might be reasonable when we revisit \cref{eq:pt} and \cref{eq:sft}. The probability $P_\mathrm{LM}(X_\mathrm{PT})$ of the next token prediction for the foundation model is conditioned on the distribution of the pre-training text $X_\mathrm{PT}$. 
SFT narrows the probability space for token selection, adjusting the parameters to better align with the distribution, \textit{i.e.}, the probability $P_\mathrm{SFT}(y|X_\mathrm{ins.})$ is conditioned on the distribution of the instruction text $X_\mathrm{ins.}$.

\paragraph{Experimental Settings}
To validate the aforementioned hypothesis, we selected the representative cross-lingual task of machine translation as our analytical testbed. The main research method involved quantifying the differences and similarities in the decision space and token selection behavior between the foundation LLM and the SFT-aligned LLM. 
For the model selection, we chose the foundation Llama2 7B model and conducted supervised fine-tuning on it using the Alpaca dataset\footnote{\url{https://github.com/tatsu-lab/stanford\_alpaca}}\cite{alpaca}. 
The optimization was carried out using a cosine learning rate scheduler, with the maximum learning rate set to $2e-5$ and a warmup ratio of $0.03$. 
Training was performed on two Nvidia-H800 GPUs using LoRA parameter-efficient fine-tuning \cite{hu2021lora} technique, with a cumulative batch size of $64$. Other hyper-parameters follow those of the original Alpaca settings.

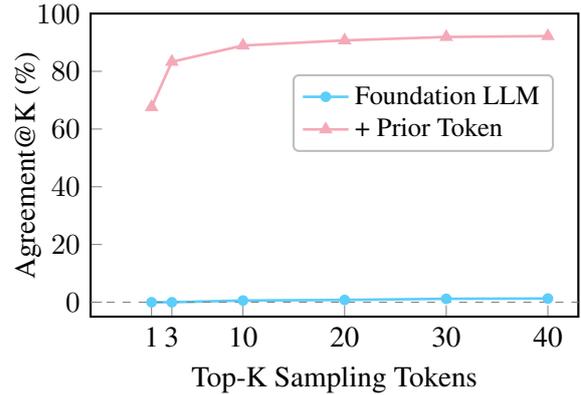
\begin{figure}
    \centering
    \begin{tikzpicture}
    \begin{axis}[
        xmin=-5, xmax=43,
        ymin=-5, ymax=100,
        xtick={1,3,10,20,30,40},
        xticklabels={1,3,10,20,30,40},
        ylabel={Agreement@K (\%)},
        xlabel={Top-K Sampling Tokens},
        ylabel style={align=center, at={(axis description cs:-0.08,0.5)},anchor=south},
        xlabel style={align=center},
        ytick={0, 20, 40, 60, 80, 100},
        legend cell align=left,
        legend style={at={(0.95,0.8), font=\fontsize{10}{12}\selectfont},
        draw=lightgray, %
        line width=0.8pt, 
        rounded corners=2pt, 
        fill=white, 
        text=black, 
        draw opacity=1, 
        },
        xtick pos=bottom,
        ytick pos=left,
        width=0.5\textwidth,
        height=0.35\textwidth,
        axis line style={line width=0.8pt}, 
        ]
        \addplot[
            color=myblue,
            mark=*,
            line width=1pt, 
            mark size=1.5pt,
            ]
            coordinates {
        (1, 0.0)
         (3, 0.0)
         (10, 0.592885375494071)
         (20, 0.7905138339920948)
         (30, 1.185770750988142)
         (40, 1.2845849802371543)
            };
            \addlegendentry{Foundation LLM}
        \addplot[
            color=mypink,
            mark=triangle*,
            line width=1pt, 
            mark size=2pt,
            ]
            coordinates {
             (1, 67.58893280632411)
             (3, 83.300395256917)
             (10, 88.93280632411067)
             (20, 90.71146245059289)
             (30, 91.89723320158103)
             (40, 92.19367588932806)
            };
            \addlegendentry{+ Prior Token}
            \addplot[dashed, color=gray] coordinates {(-5, 0) (43, 0)};
      \end{axis}
    \end{tikzpicture}
    \caption{The agreement between the SFT model and the foundation model in terms of the selection of the next token. Once the Prior Token is provided, the token chosen by the SFT model is also can be found within the Top-K candidate words of foundation model.}
    \label{fig:token-coverage}
\end{figure}
\begin{figure*}
    \centering
    \includegraphics[width=\textwidth]{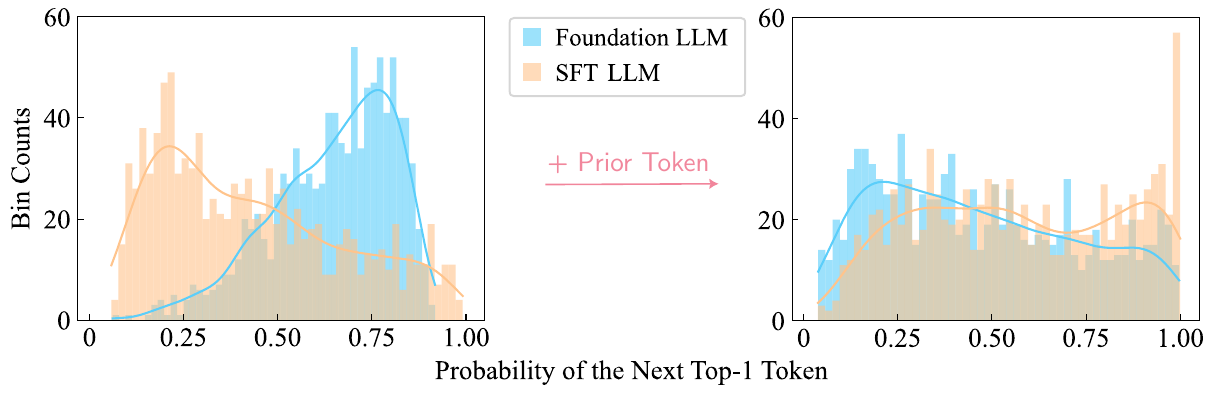}
    \caption{The probability distribution of tokens selected by various models. Incorporation of a Prior Token causes the decision probabilities of both models to converge across all data instances.}
    \label{fig:distri-dynamics}
\end{figure*}
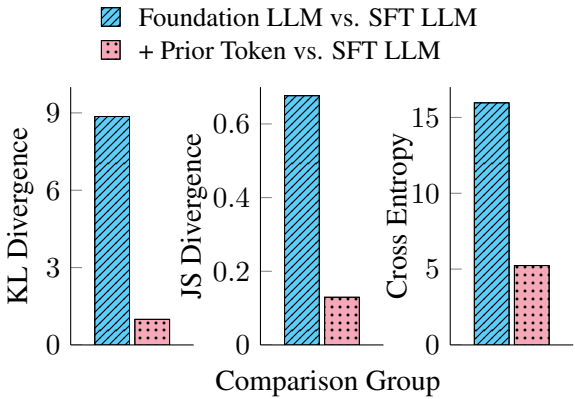
\begin{figure}
\pgfplotsset{width=3.2cm, height=5cm,
    /pgfplots/ybar legend/.style={
    /pgfplots/legend image code/.code={%
       \draw[##1,/tikz/.cd,yshift=-0.25em]
        (0cm,0cm) rectangle (7pt,7pt);},
   },}
    \centering
    \begin{tikzpicture}  
    \begin{groupplot}[
          group style={
          group name=plot,
          horizontal sep=25pt,
          vertical sep=0pt,
          group size=3 by 1
          },
          ]
      \nextgroupplot[
            ybar,
            ymin=0, ymax=10,
            ytick={0, 3, 6, 9},
            major x tick style = transparent,
            bar width=13pt,
            enlarge x limits=0.25,
            ylabel={KL Divergence},
            symbolic x coords={KL Divergence},  
            xtick=data,
            xticklabels={},
          xlabel={Comparison Group},
          xlabel style = {
                at={(ticklabel cs:0)},
                anchor=east,
                xshift=5cm,
                yshift=-0.1cm
            },
            y label style={at={(axis description cs:-0.22,0.5)},anchor=south},
                legend style={draw=none},
                axis x line*=bottom,
                axis y line*=left,
        legend cell align=left,
                legend style={
                        at={(1.8,1.05)},
                        anchor=south,
                        column sep=1ex,
                        font=\fontsize{10}{12}\selectfont,
                }
            ]  
        \addplot[ybar, draw=black, line width=0.1pt,  fill=myblue, postaction={pattern=north east lines}] coordinates {
            (KL Divergence, 8.859531)
        };  
        \addplot[ybar, draw=black, line width=0.1pt,  fill=mypink,  postaction={pattern=dots}] coordinates {
            (KL Divergence, 0.99212813)
        };  
        \legend{
            Foundation LLM vs. SFT LLM, 
            + Prior Token vs. SFT LLM, 
            }
        ]
      \nextgroupplot[
            ybar,
            ymin=0, ymax=0.7,
            ytick={0, 0.2, 0.4, 0.6},
            major x tick style = transparent,
            bar width=13pt,
            enlarge x limits=0.25,
            ylabel={JS Divergence},
            y label style={at={(axis description cs:-0.35,0.5)},anchor=south},
            symbolic x coords={JS Divergence},  
            xtick=data,
            xticklabels={}, 
                axis x line*=bottom,
                axis y line*=left,
        legend cell align=left,
                legend style={
                        at={(1,1.05)},
                        anchor=south east,
                        column sep=1ex,
                        font=\fontsize{10}{12}\selectfont,
                }
            ]  
        \addplot[ybar, draw=black, line width=0.1pt,  fill=myblue,  postaction={pattern=north east lines}] coordinates {
            (JS Divergence, 0.676708801874059)
        };
        \addplot[ybar, draw=black, line width=0.1pt,  fill=mypink,  postaction={pattern=dots}] coordinates {
            (JS Divergence, 0.12943371464206008)
        }; 
        ]
      \nextgroupplot[
            ybar,
            ymin=0, ymax=17,
            ytick={0, 5, 10, 15},
            major x tick style = transparent,
            bar width=13pt,
            enlarge x limits=0.25,
            symbolic x coords={Cross Entropy},
            ylabel={Cross Entropy},
            xticklabels={}, 
            y label style={at={(axis description cs:-0.22,0.5)},anchor=south},
            xtick=data,  
                axis x line*=bottom,
                axis y line*=left,
        legend cell align=left,
                legend style={
                        at={(1,1.05)},
                        anchor=south east,
                        column sep=1ex,
                        font=\fontsize{10}{12}\selectfont,
                }
            ]  
        \addplot[ybar, draw=black, line width=0.1pt,  fill=myblue,  postaction={pattern=north east lines}] coordinates {
            (Cross Entropy, 15.964761347945098)
        };
        \addplot[ybar, draw=black, line width=0.1pt,  fill=mypink,  postaction={pattern=dots}] coordinates {
            (Cross Entropy, 5.234162323806589)
        }; 
        ]
    \end{groupplot}
    \end{tikzpicture}  
    \caption{The divergence in probability distributions across the entire vocabulary during decoding. Prior Token significantly reduces the discrepancy between the foundation model and the SFT model.}
    \label{fig:bar-div-analysis}
\end{figure}

\paragraph{A Prior Token Elicits Silent Majority}
Inspired by the categorization of token shifts by \citet{lin2023unlocking}, we propose to quantify the agreement of token selection between foundation LLM $\theta_\mathrm{PT}$ and SFT LLM $\theta_\mathrm{SFT}$. 
Given the same prefix input $\hat{X}$, we aim to measure whether the next token selected by the SFT LLM, $y_{\mathrm{SFT}}$, is among the top-$K$ tokens, $\mathbf{y}_{\mathrm{PT}}$, with the highest probabilities in the decision space of the foundation LLM, which can be formally expressed as follows:

\begin{gather}
   y_\mathrm{SFT} = \myargmax\limits_{y \in V}~{P(y|\hat{X};\theta_\mathrm{SFT})} \notag \\
   \mathbf{y}_\mathrm{PT} = \{ y | \mathop{\mathrm{arg\,top}K}\limits_{y \in V} P(y|\hat{X};\theta_\mathrm{PT}) \} \notag \\
    \mathrm{Aggrement}_{K} = \frac{1}{L}\sum_{l=1}^{L}\mathds{1}_{y_\mathrm{SFT} \in \mathbf{y}_\mathrm{PT}  }
\end{gather}
where $V$ is the vocabulary shared by two models, and $L$ is the length of the dataset.

We compare the agreement of the token selection made by the models under the \textbf{same} prefix text $\hat{X}$ in two different experimental setups. The first setup uses the instruction text as the prefix, \textit{i.e.}, $\hat{X} = X_\mathrm{ins.}$; the second takes the first token decoded by the SFT model as a prior token, appending it to the original instruction prefix, \textit{i.e.}, $\hat{X} = \left[X_\mathrm{ins.}, y_\mathrm{SFT}^{(1)}\right]$. 
For the SFT model, the second setup is equivalent to continuing its own decoding behavior, whereas for the foundation model, it becomes decoding with the addition of a prior token.

\cref{fig:token-coverage} illustrates the agreement between the foundation model's predictions and those of the SFT model regarding the selection of the next token, given an identical text prefix. Across the entire translation data, it is observed that after incorporating merely one prior token, the foundation model exhibits a high degree of agreement with the SFT model in terms of token selection. This demonstrates that the alignment effect of SFT in cross-lingual generation tasks is also somewhat superficial.
Even in instances where the token with the highest probability differs between the two models, 90.8\% of the tokens chosen by the SFT model are present within the ``\textit{silent majority}'' in the decision space of the foundation model, specifically, among the top 20 most probable token choices.

\paragraph{Lens of Distribution}
Instead of focusing on the coverage of token selection outcomes, we also observe the decision dynamics and similarities from the perspective of the overall probability distribution, with the data settings consistent with the previous setup.
First, as shown in \cref{fig:distri-dynamics}, after adding a prior token, the probability of the next tokens chosen by both models have closely aligned distributions. 
The reason that the foundation model exhibits a high probability given the instruction text as a prefix lies in a preference for choosing to continue the instruction text rather than completing the cross-linguistic semantic transformation.
Additionally, we quantify the distribution disparities between the two models through the probability distribution of the vocabulary.
The disparity metrics used include Kullback-Leibler (KL) divergence, Jensen-Shannon (JS) divergence, and cross-entropy \cite{kullback1997information}. As depicted in \cref{fig:bar-div-analysis}, the disparity of decision space of the foundation model significantly decreases after adding the prior token, aligning more closely with the SFT model.

These findings indicate that such prior tokens serve a dual function: they not only steer the foundation model towards generating tokens pertinent to cross-lingual generation but also modulate the decision space to align more closely with the task-specific distribution.

\section{Pretty: Prefix TexT as a Yarn}
\subsection{Motivation}
The observations discussed earlier confirm that SFT effectively narrows the decision space of the foundation model during text generation that is conditioned on instruction text. 
The disparity in token selection between the foundation LLM and the SFT LLM, however, might not be reduced by a training-based transfer methodology. 
By appending a prior token into the instruction text, the choices of the next token between the two models tend to become largely consistent, and in the vast majority of cases, the tokens chosen by SFT model are also found within the high-probability candidate words of foundation model. 
These phenomena show that the alignment elicited by SFT is somewhat superficial in cross-lingual generation tasks and motivate us to propose a training-free alignment method by leveraging these prior tokens.

\subsection{Formulation}
Upon revisiting \cref{eq:pt} and \cref{eq:sft}, the goal of proposing a training-free approach is to enable the conditional decoding probability of foundation model to approximate those of SFT model. 
Therefore, ideally, the selected prior tokens $X_{\mathrm{pri.}} = \{x_{\mathrm{pri.}}\}$ may satisfy the following criteria:
\begin{align}\label{eq:align}
        P(\mathbf{y}_{\mathrm{PT}}|\left[X_{\mathrm{ins.}}, X_{\mathrm{pri.}}\right];\theta_\mathrm{PT}) \notag \\ \approx P(\mathbf{y}_{\mathrm{SFT}}|X_{\mathrm{ins.}};\theta_\mathrm{SFT})
\end{align}
where $\mathbf{y}_\mathrm{PT}$ and $\mathbf{y}_\mathrm{SFT}$ represent the outputs of the foundation and the SFT models, respectively. It is important to note that a single prior token may not serve as an optimal solution due to its non-derivable characteristic. Hence, we extend our methodological approach to include appending multiple prior tokens, grouping them to form a prefix text.

\subsection{Construction of Prior Tokens}
To ensure that the proposed method is applicable to a wide array of languages, we propose three construction strategies based on the availability of language resources, aiming to guarantee the universality of our approach.

\paragraph{SFT Prior} represents an ideal scenario where the first few tokens generated by a SFT model are used as priors. This method is theoretically rational when the SFT model is derived from the same foundation model because it directly approximates \cref{eq:align} by sampling $x_{\mathrm{pri.}} \sim \{y_{\mathrm{SFT}}\}$. In practical applications, this might be suitable for high-resource languages due to the imbalanced language capabilities of other languages. 
Additionally, SFT could potentially degrade the knowledge and abilities that the foundation model has already acquired. In such cases, using prior tokens from the SFT model can contribute to generating better results. This situation will be discussed further in the subsequent section.

\paragraph{Refined Prior} is more readily accessible for most languages and tasks. We can utilize the output tokens generated by a smaller model trained for specific downstream tasks and use them as prior tokens to achieve weak-to-strong generalization \cite{burns2023weak}.

\paragraph{Pseudo Prior}
For extremely low-resource language pairs, where there is no labeled data for downstream tasks, both SFT and Refined priors are difficult to obtain. 
For cross-lingual tasks, we can create pseudo labels in target language as prior tokens. 
For instance, in machine translation tasks, we might use bilingual dictionaries to acquire pseudo prior tokens. 
However, the quality and accuracy of pseudo labels remain uncertain, and the extent of their impact on the generative performance of the foundation LLM is not yet clear. We will explore this problem further in the context of experimental results discussed later in the paper.

\section{Experiments}
We examine the effectiveness of our proposed training-free alignment method on two distinct tasks: machine translation, cross-lingual summarization and non-English POS tagging. 
Machine translation serves as a prototypical cross-lingual generation task, entailing the transformation of a sequence from a source language to a target language \cite{bahdanau2014neural, vaswani2017attention, DBLP:conf/acl/Zhan0WZCZ23}. 
{As for cross-lingual summarization, it requires the model to generate a summary of an article in a different language \cite{bhattacharjee-etal-2023-crosssum, chen-etal-2023-revisiting}.} 
Although POS tagging \cite{manning2011part, de2021universal, chiche2022part} primarily assesses the model's ability to understand monolingual text, we include it as multilingual experiments to show the universality of our methods.
\begin{table*}[t!]
\centering
\scalebox{0.78}{
\begin{tabular}{lccccccccccc}
\toprule
\multicolumn{12}{c}{\textit{{English-Centric}}}                                                                                                        \\
                                           \midrule
                                           \multirow{2}{*}{\textbf{\textbf{Models}}} & \multicolumn{2}{c}{\textbf{En-Zh}} & \multicolumn{2}{c}{\textbf{En-Uk}} & \multicolumn{2}{c}{\textbf{Zh-En}} & \multicolumn{2}{c}{\textbf{Uk-En}} & \multicolumn{2}{c}{\textbf{Avg.}}         & {\textbf{\%SFT.}} \\
\cmidrule(lr){2-3} \cmidrule(lr){4-5} \cmidrule(lr){6-7} \cmidrule(lr){8-9}  \cmidrule(lr){10-11} \cmidrule(r{1mm}){12-12}
                                           & spBL.       & CoM.      & spBL.       & CoM.      & spBL.       & CoM.      & spBL.       & CoM.      & spBL. & CoM. & All \\
\midrule 
\rowcolor{gray!20} \multicolumn{12}{c}{\textit{{Llama2-7B}}} \\
Llama2-7B-Alpaca                                                     & 13.6                 & \underline{80.9}                 & 24.0                 & 83.3                 & 23.5                 & \underline{85.1}                 & 34.4                 & 85.5                 & 23.9          & 83.7             & -                    \\
Llama2-7B-Chat                                                       & 7.8                  & 67.2                 & 18.1                 & 71.0                 & 18.5                 & 81.3                 & 30.4                 & 83.3                 & 18.7            & 75.7           & -               \\
\hdashline\noalign{\vskip 0.5ex}
Llama2-7B$_{\textsc{prompting}}$                                     & 5.9                  & 64.1                 & 11.0                 & 60.9                 & \underline{\textbf{24.3}}        & 84.8                 & 34.2                 & 85.0                 & 18.9               & 73.7        & 80.4               \\
Llama2-7B                                                            & 7.7                  & 72.0                 & 0.2                  & 32.4                 & 12.0                 & 74.4                 & 9.3                  & 59.2                 & 7.3              & 59.5          & 52.5               \\
\hdashline\noalign{\vskip 0.5ex}
\multicolumn{1}{r}{+\prettytext~(SFT Prior)}     & 13.3                 & 80.0                 & 23.0                 & 83.1                 & 23.7                 & {84.9}        & 33.6                 & 85.3                 & 23.4                & 83.3       & 98.8               \\
\multicolumn{1}{r}{+\prettytext~(Pseudo Prior)}  & 12.0                 & 75.7                 & 18.1                 & 74.1                 & 16.9                 & 80.3                 & 27.2                 & 78.3                 & 18.6               & 77.1        & 85.4               \\
\multicolumn{1}{r}{+\prettytext~(Refined Prior)} & \underline{\textbf{14.2}}        & \textbf{80.5}        & \underline{\textbf{24.1}}        & \underline{\textbf{83.8}}        & 24.0                 & \textbf{84.9}        & \underline{\textbf{34.6}}        & \underline{\textbf{85.6}}        & \underline{\textbf{24.2}}         & \underline{\textbf{83.7}}     & \textbf{100.9}     \\
\rowcolor{gray!20} \multicolumn{12}{c}{\textit{{Mistral-7B}}} \\
Mistral-7B-Instruct      & 6.6                 & 64.6                & 20.3                & 78.2                & 20.5                & \underline{83.2}    & \underline{32.9}    & \underline{84.8}    & 20.1                & 77.7                & -                \\
\hdashline\noalign{\vskip 0.5ex}
Mistral-7B & 1.2                 & 42.6                & 0.3                 & 30.8                & 19.9                & 77.1          & 21.5          & 69.4          & 10.7                & 55.0                & 46.2           \\
\hdashline\noalign{\vskip 0.5ex}
\multicolumn{1}{r}{+\prettytext~(SFT Prior)}     & 13.8                & 78.1                & 23.1                & 79.2                & 20.0                & 82.3          & 32.1          & 83.3          & 22.3                & 80.7                & 117.2          \\
\multicolumn{1}{r}{+\prettytext~(Pseudo Prior)}  & 13.3                & 75.8                & 20.1                & 75.7                & 16.5                & 79.7          & 24.9          & 77.3          & 18.7                & 77.1                & 107.2           \\
\multicolumn{1}{r}{+\prettytext~(Refined Prior)}  & \underline{\textbf{15.9}} & \underline{\textbf{81.3}} & \underline{\textbf{24.9}} & \underline{\textbf{82.9}} & \underline{\textbf{21.5}} & \textbf{83.0} & \textbf{32.3} & \textbf{83.9} & \underline{\textbf{23.7}} & \underline {\textbf{82.7}} & \textbf{124.6}    \\
\midrule
\multicolumn{12}{c}{\textit{{Non-English-Centric}}}                      \\
                                           \midrule
                                           \multirow{2}{*}{\textbf{\textbf{Models}}} 
                                           & \multicolumn{2}{c}{\textbf{De-Fr}} & \multicolumn{2}{c}{\textbf{Fr-De}} & \multicolumn{2}{c}{\textbf{Zh-Pt}} & \multicolumn{2}{c}{\textbf{Pt-Zh}} & \multicolumn{2}{c}{\textbf{Avg.}}         & {\textbf{\%SFT.}} \\
\cmidrule(lr){2-3} \cmidrule(lr){4-5} \cmidrule(lr){6-7} \cmidrule(lr){8-9} \cmidrule(r){10-11} \cmidrule(r{1mm}){12-12}
                                           & spBL.       & CoM.      & spBL.       & CoM.      & spBL.       & CoM.      & spBL.       & CoM.      & spBL. & CoM. & All        \\
\midrule
\rowcolor{gray!20} \multicolumn{12}{c}{\textit{{Llama2-7B}}} \\
Llama2-7B-Alpaca                                                     & 29.8                 & \underline{81.5}                 & \underline{24.1}                 & \underline{80.9}                 & 16.6                 & \underline{81.4}                 & 11.3                 & 78.6                 & \underline{20.5}                & 80.6       & -                    \\
Llama2-7B-Chat                                                       & 6.2                  & 68.0                 & 7.3                  & 64.5                 & 3.0                  & 67.8                 & 6.2                  & 66.6                 & 5.7             & 66.7           & -                    \\
\hdashline\noalign{\vskip 0.5ex}
Llama2-7B$_{\textsc{prompting}}$                                     & 22.2                 & 77.4                 & 15.4                 & 73.3                 & 14.4                 & 78.9                 & 4.4                  & 64.1                 & 14.1          & 73.4             & 78.5              \\
Llama2-7B                                                            & 1.0                  & 51.1                 & 3.2                  & 54.0                 & 0.9                  & 61.4                 & 7.3                  & 70.0                 & 3.1            & 59.1            & 47.6              \\
\hdashline\noalign{\vskip 0.5ex}
\multicolumn{1}{r}{+\prettytext~(SFT Prior)}     & 28.2                 & 80.6                 & 23.0                 & \textbf{80.4}        & 16.3                 & 81.1                 & 10.5                 & 77.4                 & 19.5                & 79.9      & 97.2             \\
\multicolumn{1}{r}{+\prettytext~(Pseudo Prior)} &    18.3 & 68.9 &	17.3 & 72.2 & 11.6 &70.4 & 5.0	& 65.6 & 13.1 & 69.3 & 73.9\\
\multicolumn{1}{r}{+\prettytext~(Refined Prior)} & \underline{\textbf{29.1}}        & \textbf{81.4}        & \textbf{22.9}        & \textbf{80.4}        & \underline{\textbf{17.1}}        & \textbf{81.1}        & \underline{\textbf{12.2}}        & \underline{\textbf{79.4}}        & \textbf{20.3}        & \underline{\textbf{80.6}}  & \textbf{100.4}    \\

\rowcolor{gray!20} \multicolumn{12}{c}{\textit{{Mistral-7B}}} \\
Mistral-7B-Instruct      & 22.1                & 76.1                & 20.4                & 75.9                & 10.5                & 74.8                & 3.3                 & 60.2                & 14.1          & 71.8          & -                \\
\hdashline\noalign{\vskip 0.5ex}
Mistral-7B & 1.2                 & 46.1                & 1.6                 & 40.6                & 1.0                 & 52.8                & 0.4                 & 43.6                & 1.1           & 45.8          & 36.5           \\
\hdashline\noalign{\vskip 0.5ex}
\multicolumn{1}{r}{+\prettytext~(SFT Prior)}     & 20.1                & 73.3                & 20.7                & 75.1                & 11.0                & 74.7                & 6.8                 & 67.3                & 14.7          & 72.6          & 113.8          \\
\multicolumn{1}{r}{+\prettytext~(Pseudo Prior)}  & 18.1                & 66.4                & 17.3                & 70.4                & 5.9                 & 65.6                & 3.7                 & 59.4                & 11.3          & 65.5          & 87.7           \\
\multicolumn{1}{r}{+\prettytext~(Refined Prior)}  & \underline{\textbf{28.3}} & \underline{\textbf{78.8}} & \underline{\textbf{22.3}} & \underline{\textbf{78.5}} & \underline{\textbf{14.2}} & \underline{\textbf{78.6}} & \underline{\textbf{13.6}} & \underline{\textbf{80.6}} & \underline{\textbf{19.6}} & \underline{\textbf{79.1}} & \textbf{153.8} \\
\bottomrule
\end{tabular}
}
\caption{Translation performance of different models on Flores-101 subsets. \textbf{Bold} values indicate that the best performance among foundation models. The overall best results are \underline{underlined}. ``\%SFT.'' denotes the relative performance compared to the best SFT model of each family. }
\label{tab:mt-all}
\end{table*}

\subsection{Experimental Settings} 
\paragraph{Data}
{We use Flores-101 \cite{goyal2022flores}, CrossSum \cite{bhattacharjee-etal-2023-crosssum} as benchmarks for machine translation and cross-lingual summarization tasks, respectively.}
For POS tagging tasks, we choose the POS test split from the XGLUE benchmark \cite{liang-etal-2020-xglue}, which is derived from the Universal Dependencies Treebank v2.5.
To investigate the performance across various resource languages, we carefully selected eight languages based on the pre-training data proportions disclosed in the Llama2 technical report \cite{touvron2023llama}. 
These languages are French, German, Chinese, Russian, Ukrainian, Portuguese, Hindi and Arabic. Among these, the first four languages account for more than 0.1\% of the pre-training data of Llama2, while Ukrainian and Portuguese fall below 0.1\%, Hindi and Arabic is below 0.05\%. For the Llama2 model, we can categorize these three types of languages into high-resource languages, low-resource languages, and extremely low-resource languages, respectively.

\paragraph{Models and Baselines} 
The settings of Llama2 foundation model and the SFT model are consistent with those described in \cref{sec:prelimi}.
{To further demonstrate the generality of our proposed method, we incorporated the Mistral-7B LLM family \cite{mistral23} into our experiments, covering both out-of-the-box SFT and foundation models.}

In the machine translation task, the Llama2 foundation model does not tend to generate translations when given explicit translation instructions. While this is a normal phenomenon according to our previous discussion, to ensure a fair comparison, we also searched for a better prompts for the foundation model. This prompting approach is referred to as ``Llama2-7B$_{\textsc{Prompting}}$'' in subsequent sections. 
For POS tagging, we experimented with various instructions and selected one that consistently prompts both the foundation model and the SFT model to reliably generate classification results in text.
{Although we report the zero-shot performance for the aforementioned tasks, we found that even out-of-the-box SFT models cannot produce stable output for cross-lingual summarization task. Hence, we prepend a constant demonstration before the input to also assess the effectiveness of our proposed method under the in-context learning paradigm \cite{dong2023survey}.}

\paragraph{Sources of Prior Token}
The sources of crafting prior tokens include:
\begin{itemize}
    \item \textbf{SFT Prior}: We took the first $k$ tokens of output produced by SFT model as the prior tokens. {For multiple SFT models, we select the model that demonstrates better performance.} 
    \item \textbf{Refined Prior}: We use downstream task models with smaller parameter sizes as the source of refined priors. For the different tasks, we utilized the distilled 600M variant of NLLB-200 translation model\footnote{\url{https://huggingface.co/facebook/nllb-200-distilled-600M}}\cite{costa2022no}, mT5 cross-lingual summarization model\footnote{\url{{https://hf.co/csebuetnlp/mT5_m2m_crossSum}}} and the Unicoder-NLU model\footnote{\url{https://github.com/microsoft/Unicoder/}}\cite{huang2019unicoder}, respectively. 
    \item \textbf{Pseudo Prior}: The pseudo prior is applied to two cross-lingual tasks since it can utilize cross-lingual language resources. 
    We create pseudo prior tokens for machine translation task by referencing dictionary \footnote{Please refer to \cref{sec:dic} for dictionary information.} entries.
    For cross-lingual summarization, we initially extract keywords from each passage using KeyBERT \cite{grootendorst2020keybert} and then perform word-by-word translation.
    However, not all initial sentence tokens will be covered by the dictionary. To handle such instances, a back-off strategy is implemented, where the target language equivalent of the first available dictionary token is used as the prior token.

\end{itemize}
For two cross-lingual task, the first $k=2$ tokens are chosen as the prior tokens. This helps to avoid inadequate guidance from single non-informative tokens like punctuation or numbers. In the case of the pseudo prior, due to the back-off strategy, only one token is used for fair comparison.
For POS tagging task, the strategy is more straightforward with only the first $k=1$ label considered as the prior token.

\begin{table*}[t!]
\centering
\scalebox{0.72}{
\begin{tabular}{lcccccccccccccccc}
\toprule
                                           \multirow{2}{*}{\textbf{\textbf{Models}}} & \multicolumn{3}{c}{\textbf{En-Zh}} & \multicolumn{3}{c}{\textbf{En-Hi}} & \multicolumn{3}{c}{\textbf{Uk-Pt}} & \multicolumn{3}{c}{\textbf{Ar-Ru}} & \multicolumn{3}{c}{\textbf{Avg.}}         & {\textbf{\%SFT.}} \\
\cmidrule(lr){2-4} \cmidrule(lr){5-7} \cmidrule(lr){8-10} \cmidrule(lr){11-13}  \cmidrule(lr){14-16} \cmidrule(r{1mm}){17-17}
& R2       & RL      & LS       & R2       & RL      & LS & R2       & RL      & LS & R2       & RL      & LS & R2       & RL      & LS & All \\
\midrule 
\rowcolor{gray!20} \multicolumn{17}{c}{\textit{{Llama2-7B} w/ Constant 1-Shot Demonstration}} \\
Llama2-7B-Alpaca                                 & 7.0                  & 12.4                 & 11.9                 & \underline{1.7}                  & 10.7                 & 17.3                 & 1.5                  & 6.1                  & 5.8                  & 0.1                  & 0.5                  & 1.3                  & 2.6                  & 7.4                  & 9.1                  & -                    \\
Llama2-7B-Chat                                   & 6.3                  & 11.6                 & 8.7                  & 1.5                  & \underline{11.7}           & \underline{27.1}           & {2.5}            & 8.3                  & 7.1                  & 0.0                  & 0.3                  & 0.2                  & 2.6                  & 8.0                  & 10.7                 & -                    \\
\hdashline\noalign{\vskip 0.5ex}
Llama2-7B                                        & 9.3                  & 16.6                 & 29.2                 & \textbf{1.6}                  & 10.2                 & {15.3}        & 0.8                  & 4.0                  & 1.9                  & 0.6                  & 4.1                  & 15.5                 & 3.1                  & 7.6                  & 12.1                 & 262.4                \\
\hdashline\noalign{\vskip 0.5ex}
\multicolumn{1}{r}{+\prettytext~(SFT Prior)}     & 7.4                  & 13.9                 & 25.9                 & 1.5                  & 9.7                  & 12.9                 & 1.9                  & 6.7                  & 9.8                  & 0.1                  & 0.4                  & 0.8                  & 2.7                  & 6.7                  & 9.8                  & 106.3                \\
\multicolumn{1}{r}{+\prettytext~(Pseudo Prior)}  & 8.0                  & 14.5                 & 29.1                 & 1.4                  & 9.9                  & 14.5                 & {{2.5}}   & 9.1                  & \underline{\textbf{13.6}}                 & 1.2                  & 5.9                  & 23.5                 & 3.3                  & 8.5                  & 15.4                 & 387.5                \\
\multicolumn{1}{r}{+\prettytext~(Refined Prior)} & \underline{\textbf{11.2}}  & \underline{\textbf{19.0}}  & \underline{\textbf{32.6}}  & \textbf{1.6}         & {\textbf{10.8}}  & {\textbf{15.9}} &  \underline{\textbf{3.4}} & \underline{\textbf{10.5}}        & 11.3 & \underline{\textbf{1.5}}   & \underline{\textbf{7.9}}   & \underline{\textbf{30.1}}  & \underline{\textbf{4.4}}   & \underline{\textbf{10.5}}  & \underline{\textbf{17.5}}  & \textbf{490.6}       \\
\rowcolor{gray!20}\multicolumn{17}{c}{\textit{{Mistral-7B}~w/ Constant 1-Shot Demonstration}} \\
Mistral-7B-Instruct                              & 5.9                  & 12.2                 & 17.2                 & 1.0                  & 10.3                 & \underline{23.4}                 & 1.5                  & 6.2                  & 17.7                 & 0.4                  & 2.6                  & 12.8                 & 2.2                  & 7.8                  & 17.8                 & -                    \\
\hdashline\noalign{\vskip 0.5ex}
Mistral-7B                                       & 12.3                 & 20.9                 & 44.5                 & 1.6                  & 10.6                 & 17.6                 & 4.8                  & 12.9                 & 27.7                 & 1.8                  & 6.5                  & 23.3                 & 5.1                  & 11.2                 & 21.6                 & 206.1                \\
\multicolumn{1}{r}{+\prettytext~(SFT Prior)}     & 9.7                  & 17.6                 & 40.7                 & 1.4                  & 10.0                 & 17.0                 & 2.3                  & 7.9                  & 17.5                 & 0.2                  & 1.1                  & 3.2                  & 3.4                  & 8.0                  & 15.0                 & 114.5                \\
\multicolumn{1}{r}{+\prettytext~(Pseudo Prior)}  & 9.9                  & 17.5                 & 41.0                 & 1.4                  & 9.9                  & 17.4                 & 3.1                  & 11.6                 & 35.1                 & 1.7                  & 7.9                  & 32.9                 & 4.0                  & 10.2                 & 23.5                 & 195.8                \\
\multicolumn{1}{r}{+\prettytext~(Refined Prior)} & \underline{\textbf{15.0}}  & \underline{\textbf{24.1}}  & \underline{\textbf{49.6}}  & \underline{\textbf{1.8}}   & \underline{\textbf{11.3}}  & \textbf{19.7}        & \underline{\textbf{5.5}}   & \underline{\textbf{16.5}}  & \underline{\textbf{46.9}}  & \underline{\textbf{2.6}}  & \underline{\textbf{10.9}}  & \underline{\textbf{42.0}}  & \underline{\textbf{6.2}}   & \underline{\textbf{13.8}}  & \underline{\textbf{29.7}}  & \textbf{275.6}       \\
\bottomrule
\end{tabular}
}
\caption{Summarization performance of different models on CrossSum subsets. ``R2/L'' and ``LS'' refer to the ROUGE and LaSE score, respectively. \textbf{Bold} values indicate that the best performance among foundation models. The overall best results are \underline{underlined}. ``\%SFT.'' denotes the relative performance compared to the best SFT model. }
\label{tab:cs-all}
\end{table*}

\begin{table*}[t!]
\centering
\scalebox{0.85}{
\begin{tabular}{lcccccccccccc}
\toprule
\multirow{2}{*}{\textbf{Models}} & \multicolumn{2}{c}{\textbf{Fr}} & \multicolumn{2}{c}{\textbf{Zh}} & \multicolumn{2}{c}{\textbf{Pt}} & \multicolumn{2}{c}{\textbf{Ru}} & \multicolumn{2}{c}{\textbf{Ar}} & {\textbf{Avg.}} & {\textbf{\%SFT.}} \\
\cmidrule(lr){2-3} \cmidrule(lr){4-5} \cmidrule(lr){6-7} \cmidrule(lr){8-9} \cmidrule(lr){10-11} \cmidrule(r){12-12} \cmidrule(r){13-13}
& Prec. & F$_1$ & Prec. & F$_1$ & Prec. & F$_1$ & Prec. & F$_1$ & Prec. & F$_1$ & Prec. & All \\
\midrule
Llama2-7B-Alpaca                                                      & 48.2                       & 42.8  & 38.6                 & 36.3                 & 40.7                 & 35.9                 & 42.3                 & 36.7                 & 34.4                 & 30.8    &38.7 &-    \\                             
\hdashline\noalign{\vskip 0.5ex}
Llama2-7B                                                          & 45.0                        & 37.9                 & 39.8                  & 36.2                 & 39.8                  & 33.2                 & 42.5                 & 33.8                 & 36.5                  & 32.1  &37.7   &97.4 \\              
\hdashline\noalign{\vskip 0.5ex}
\multicolumn{1}{r}{+\prettytext~(SFT Prior)}   & 54.8                       & 50.0    & 38.0                 & 33.5                 & 49.1 & 45.3                 & 49.7                 & 44.1        & 35.1                 & 31.1           &43.1      &111     \\           
\multicolumn{1}{r}{+\prettytext~(Refined Prior)} & \underline{\textbf{59.3}}              & \underline{\textbf{54.8}}     & \underline{\textbf{43.0}}        & \underline{\textbf{38.8}}        & \underline{\textbf{54.5}}        & \underline{\textbf{50.6}}          & \underline{\textbf{55.3}}                  & \underline{\textbf{49.2}}        & \underline{\textbf{44.0}}        & \underline{\textbf{39.6}}  & \underline{\textbf{48.9}}   & \textbf{126}   \\
\bottomrule
\end{tabular}
}
\caption{POS tagging performance of different Llama2 models on XGLUE subsets. \textbf{Bold} values indicate that the best performance among foundation models.  The overall best results are \underline{underlined}. ``\%SFT.'' denotes the relative performance compared to Alpaca model.}
\label{tab:pos-all}
\end{table*}

\subsection{Evaluation}
To ensure the integrity of the output data from all models, we standardized the output by cleaning it in accordance with the specific output style of each model. Subsequently, we conducted a manual inspection to guarantee that only the required labels were retained.

\paragraph{Task-specific Metrics} 
We use two metrics to evaluate the performance of translation quality: spBLEU\footnote{\url{https://github.com/mjpost/sacrebleu/}} \cite{goyal2022flores} and COMET\footnote{\url{https://github.com/Unbabel/COMET}}\cite{rei2020comet}.
We employed the ROUGE \cite{lin-2004-rouge} and LaSE \cite{bhattacharjee-etal-2023-crosssum} metrics for the evaluation of summarization quality. 
For the POS tagging task, we report both the precision score and F$_1$ score.

\paragraph{Relative Performance}
We further compute the ratio of the performance scores of the foundation model to the scores of the SFT model with the application of different strategies.
This ratio serves as a metric for assessing the extent to which the foundation model approximates the SFT model's performance when different strategies are applied.

\subsection{Main Results}
\paragraph{Machine Translation}
As shown in \cref{tab:mt-all}, for the machine translation task, we use up to two prior tokens as decoding guidance, allowing the base model to achieve performance comparable to that of a model after SFT. 
Moreover, in some language pairs, the translation performance outperforms SFT model when guided by Refined Prior tokens from a smaller model.
For Llama2 model family, the prior tokens provided by the SFT model, although slightly less effective, still allow the foundation model to achieve 98\% of the performance of SFT model. On the other hand, the use of pseudo labels derived from a dictionary exhibits the least effectiveness, yet this strategy still surpasses the results achieved through costly prompt engineering.

\paragraph{Cross-lingual Summarization}
The results presented in Table \ref{tab:cs-all} indicate that the foundation model exhibited superior performance compared to the SFT model in this in-context learning scenario. For prior-guided decoding, the performance of the foundation model was degraded when using prefix tokens from the SFT model, and the small performance gap in this setting suggests that the alignment achieved by the SFT model is relatively ``superficial''. Notably, the performance of Llama2 foundation model significantly improved when other priors were provided, even when using translated keywords as pseudo labels.

\paragraph{Non-English POS tagging}
The performance results of POS tagging task are presented in \cref{tab:pos-all}. These results align with the insights gleaned from the machine translation task, specifically regarding the strategy of prior token construction. 
Notably, for POS tagging task, the performance of SFT model on most language pairs falls short of the foundation model, suggesting that SFT detrimentally affect the knowledge learned at the pre-training stage.
Encouragingly, when the foundation model empowered by auxiliary prior token surpasses the performance of SFT model as well as the prompting results of itself, highlighting the potential of our proposed method in mitigating the catastrophic forgetting problem associated with SFT.

\section{Analysis and Discussion}
\subsection{Quality of Prior Tokens}
To investigate the quality of prior tokens from different sources and how they impact the final performance, we further analyze why the prior tokens given by the SFT model are less effective than those from external auxiliary models in POS tagging task. 
Unlike the machine translation task, the positional result for the POS task is definite, so we are able to verify whether it corresponds to a ground truth label. 
The results in \cref{tab:qual_prior} confirm two points. First, even if the prior tokens provided by the SFT model are of low quality, the foundation model does not suffer from severe error propagation. Secondly, the final performance of proposed method is still associated with the quality of prior tokens. This suggests that prior tokens closely aligned with the ground truth can steer the foundation model towards a more accurate decision trajectory, thereby yielding superior performance.

\begin{table}[H]
\scalebox{0.9}{
\begin{tabular}{lccccc}
\toprule
                       & \multicolumn{1}{l}{\textbf{Fr}} & \multicolumn{1}{l}{\textbf{Zh}} & \multicolumn{1}{l}{\textbf{Pt}}        & \multicolumn{1}{l}{\textbf{Ru}} & \multicolumn{1}{l}{\textbf{Ar}}       \\               
\midrule
SFT Prior     & 18.3                            & 18.3                            & 3.74                                   & 16.3                            & 12.1                                  \\
Refined Prior & \textbf{88.9}                   & \textbf{88.9}                   & \textbf{88.54} & \textbf{87.7}                   & \textbf{79.6} \\
\bottomrule
\end{tabular}
}
\caption{Accuracy of prior tokens used in POS tagging task. SFT prior tokens are of inferior quality.}
\label{tab:qual_prior}
\end{table}

\subsection{Choice of Prior Tokens}
Based on the findings from the previous section, if incorrect labels used as prior tokens can still elicit the ability of foundation model, then could random prior tokens in the target language trigger cross-lingual generative capabilities? 
To investigate this, we attempted to use random tokens of different parts of speech as the prior tokens in the English-Chinese machine translation task.
For instance, ``Modal Prior'' refers to the use of randomly picked modal verb in Chinese as the initial token.
 
The results shown in \cref{tab:random_token} indicate that the model could not be aligned to a better decision trajectory by these random prior tokens, whether they were function words or tokens with actual meaning. This supports the validity of our proposed methods for constructing prior tokens and also supplements previous findings. From this, we can summarize some rules about prior tokens: they can be of low quality but should not be completely unrelated to the target sequence.

\begin{table}[H]
\scalebox{0.9}{
\begin{tabular}{lccc}
\toprule
                 & \textbf{spBLEU} & \textbf{COMET} & \textbf{BLEU}  \\
                 \midrule
Llama2-7B        & 7.7    & \textbf{72.01} & \textbf{16.1} \\
\hdashline\noalign{\vskip 0.5ex}
~~+ Modal Prior  & \textbf{8.0}      & 68.29  & 16.0   \\
~~+ Adverb Prior & 6.4    & 63.72 & 13.1 \\
~~+ Random Prior  & 6.2    & 57.11 & 11.5 \\
\bottomrule
\end{tabular}
}
\caption{Comparison of translation performance using three types of random prior tokens.}
\label{tab:random_token}
\end{table}

\subsection{Number of Prior Tokens}
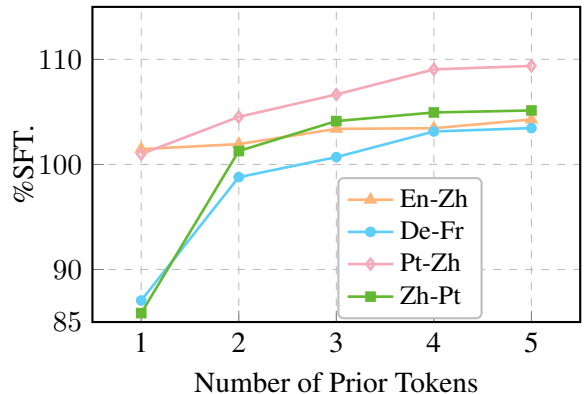
\begin{figure}[H]
    \centering
    \begin{tikzpicture}
    \begin{axis}[
        xmin=0.5, xmax=5.5,
        ymin=85, ymax=115,
        xtick={1,2,3,4,5},
        xticklabels={1,2,3,4,5},
        ylabel={\%SFT.},
        xlabel={Number of Prior Tokens},
        ylabel style={align=center, at={(axis description cs:-0.1,0.5)},anchor=south},
        xlabel style={align=center},
        ytick={85, 90, 100, 110},
        legend cell align=left,
        legend style={at={(0.8,0.46), font=\fontsize{10}{12}\selectfont},
        draw=lightgray, %
        line width=0.8pt, %
        rounded corners=2pt, 
        fill=white, %
        text=black, %
        draw opacity=1, 
        },
        xtick pos=bottom,
        ytick pos=left,
        width=0.5\textwidth,
        height=0.36\textwidth,
        grid=both, 
        grid style=dashed,
        axis line style={line width=0.8pt}, %
        ]
        \addplot[
            color=myyellow,
            mark=triangle*,
            line width=1pt, %
            mark size=2pt,
            ]
            coordinates {
        (1, 101.46)
         (2, 101.94)
         (3, 103.39)
         (4, 103.45)
          (5, 104.28)
            };
            \addlegendentry{En-Zh}
            
        \addplot[
            color=myblue,
            mark=*,
            line width=1pt, %
            mark size=1.5pt,
            ]
            coordinates {
        (1, 87.05)
         (2, 98.79)
         (3, 100.7)
         (4, 103.14)
          (5, 103.46)
            };
            \addlegendentry{De-Fr}
        \addplot[
            color=mypink,
            mark=diamond,
            line width=1pt, %
            mark size=2pt,
            ]
            coordinates {
        (1, 100.99)
         (2, 104.53)
         (3, 106.66)
         (4, 109.05)
          (5, 109.38)
            };
            \addlegendentry{Pt-Zh}
        \addplot[
            color=ppurple,
            mark=square*,
            line width=1pt, 
            mark size=1.5pt,
            ]
            coordinates {
        (1, 85.85)
         (2, 101.28)
         (3, 104.13)
         (4, 104.95)
          (5, 105.14)
            };
            \addlegendentry{Zh-Pt}
      \end{axis}
    \end{tikzpicture}
    \caption{Impact of incrementally adding refined prior tokens on performance across Flores-101 subsets.}
            \label{fig:prefix-length}
\end{figure}

Figure \ref{fig:prefix-length} depicts the relationship between the number of preceding tokens provided and the resulting changes in translation performance. It becomes apparent that performance generally improves with the addition of more tokens. 
Additionally, we note that introducing two prior tokens appears to be a performance inflection point, which may be due to instances where the initial token is a punctuation mark or a number.

\section{Conclusions}
In this paper, we investigate and analyze the decision-making discrepancies between the foundation model and the SFT model within cross-lingual generation contexts. Drawing from our analysis, we introduce a novel cross-lingual alignment method that requires no additional training and is resource-efficient. The proposed method aligns the foundation LLM to perform comparably with the SFT model solely by utilizing prefix text as priors during generation. In the future, we aim to broaden our research to encompass additional alignment scenarios, such as those involving reinforcement learning from human feedback.

\section*{Limitations}
The primary limitations of our study stem from the scope of model validation. Our research is limited to 7B models. Future endeavors should aim to extend the validation to a broader scope of models and incorporate various parameter scales to support the universality of our findings.
Furthermore, the availability of language resources is still a practical problem, particularly for low-resource languages where access to Prior Token and Refined Token sources is limited. Despite these challenges, our experimental results indicate that Pseudo Prior tokens still exhibits promising potential. 
It is important to note, however, that the development of pseudo tags may require a dedicated investigation into the linguistic rules specific to each downstream task. This process is inherently time-intensive and resource-demanding.

\section*{Acknowledgements}
This work was supported in part by the Science and Technology Development Fund, Macau SAR (Grant Nos. FDCT/0070/2022/AMJ, FDCT/060/2022/AFJ), Ministry of Science and Technology of China (Grant No. 2022YFE0204900), National Natural Science Foundation of China (Grant No. 62261160648), the Multi-year Research Grant from the University of Macau (Grant No. MYRG-GRG2023-00006-FST-UMDF), and Tencent AI Lab Rhino-Bird Gift Fund (Grant No. EF2023-00151-FST). This work was performed in part at SICC which is supported by SKL-IOTSC, and HPCC supported by ICTO of the University of Macau.

\bibliography{custom}

\newpage
\appendix

\section{Latest Benchmark}
The composition of the pre-training corpus for Llama2 is indeed unclear as Meta AI has not disclosed specifics in their technical report \cite{touvron2023llama}. Consequently, we cannot confirm the inclusion of Flores-101 datasets in the pre-training phase.
To mitigate this uncertainty, we conducted additional experiments using the WMT23 machine translation benchmark. Since Llama2 was released on July 18, 2023, and the WMT23 News Shared Task submission\footnote{\url{https://www2.statmt.org/wmt23/translation-task.html}} deadline was July 20, 2023, with the reference translations released on October 10, 2023, it is highly unlikely that WMT23 data influenced Llama2's pre-training. 

For these experiments, we used conventional BLEU as our evaluation metric because the Flores-101 benchmark introduces a specialized BLEU score variant known as spBLEU, which diverges from the conventional BLEU by adopting the sentencepiece model \cite{kudo2018sentencepiece} for tokenization. 
As shown in \cref{tab:mt-wmt23}, the experimental results indicate a consistent trend and support similar conclusions to those obtained from previous benchmarks.

\section{Detailed Setup}\label{sec:appendix}
\subsection{Llama2 Alpaca}
To avoid the need of fine-tuning for Llama2 family, we initially considered using Llama2-Chat. However, it is important to note that Llama2-Chat is a version that has been fine-tuned using both SFT and Reinforcement Learning with Human Feedback (RLHF), as it has been aligned with human preferences for helpfulness and safety. In practice, we observed that the Llama2-Chat model sometimes overly prioritizes safety, leading to instances where it refrains from generating responses due to its strict safety protocols. Consequently, we opted for a SFT model that was fine-tuned using Alpaca dataset to ensure more consistent outputs and used it as a source of obtaining SFT priors.
\paragraph{Hyper-parameters} 
The hyper-parameter settings are presented in Table \ref{tab:alpaca_hyper}. This experiment was conducted using the DeepZero 2.
\begin{table}[H]
\scalebox{0.85}{
\begin{tabular}{ll}
\toprule
\textbf{Hyper-Parameters} & \textbf{Settings}\\
\midrule
Learning Rate         & {2e-5}                   \\
LR Scheduler          & \multicolumn{1}{l}{Cosine} \\
Epoches               & 3                          \\
Seed                  & 42                         \\
Warmup Ratio          & 0.03                       \\
Weight Decay          & 0                          \\
Cumulative Batch Size & 64                         \\
LoRA Alpha            & 64                         \\
LoRA Rank             & 128                        \\
LoRA Dropout          & 0.05                       \\
Trainable Modules         & \makecell[l]{q\_proj,v\_proj,k\_proj,o\_proj,\\gate\_proj,down\_proj,up\_proj} \\
\bottomrule
\end{tabular}
}
\caption{Settings used for fine-tuning Llama2-7B foundation model with Alpaca data.}
\label{tab:alpaca_hyper}
\end{table}

\begin{table*}[t!]
\centering
\scalebox{0.95}{
\begin{tabular}{lccccccccc}
\toprule
\multirow{2}{*}{{\textbf{Models}}} & \multicolumn{2}{c}{\textbf{En-Uk}} & \multicolumn{2}{c}{\textbf{En-Zh}} & \multicolumn{2}{c}{\textbf{En-He}} & \multicolumn{2}{c}{\textbf{Avg.}}         & {\textbf{\%SFT.}} \\
\cmidrule(lr){2-3} \cmidrule(lr){4-5} \cmidrule(lr){6-7} \cmidrule(lr){8-9}  \cmidrule(r{1mm}){10-10}
                                           & BL.       & CoM.      & BL.       & CoM.           & BL.       & CoM.      & BL. & CoM. & All \\
\midrule 
\rowcolor{gray!20} \multicolumn{10}{c}{\textit{{Llama2-7B}}} \\
Llama2-7B-Alpaca  & \underline{12.8} & \underline{75.0} & \underline{ 25.9} & \underline{ 77.6} & 2.3 & 46.3 & \underline{ 13.7} & \underline{ 53.1} & - \\
\hdashline\noalign{\vskip 0.5ex}
Llama2-7B & 0.2        & 34.8     & 10.1       & 65.3       & 0.3 & 41.2 & 3.5        & 36.2       & 59.5  \\
\hdashline\noalign{\vskip 0.5ex}
\multicolumn{1}{r}{+\prettytext~(SFT Prior)}    & 12.1       & 70.3     & 24.7       & 76.4       & 1.3 & 43.4 & 12.7       & 50.7       & 94.9              \\
\multicolumn{1}{r}{+\prettytext~(Pseudo Prior)}  & 9.4        & 63.9     & 22.8       & 72.4       & 0.7 & 40.7 & 11         & 47         & 86.8               \\
\multicolumn{1}{r}{+\prettytext~(Refined Prior)} &
  \textbf{12.5} &
  \textbf{72.3} &
  \textbf{24.7} &
  \textbf{76.8} &
  \underline{\textbf{2.3}} &
  \underline{\textbf{48.3}} &
  \textbf{13.2} &
  \textbf{52.6} &
  \textbf{98.5}
\\
\bottomrule
\end{tabular}
}
\caption{Translation performance of different models on WMT23 subsets. \textbf{Bold} values indicate that the best performance among foundation models. The overall best results are \underline{underlined}. ``\%SFT.'' denotes the relative performance compared to Alpaca model. }
\label{tab:mt-wmt23}
\end{table*}

\subsection{Prompting Strategies} 
\paragraph{Machine Translation}
\begin{itemize}
    \item \textbf{Normal}:\textit{ Translate the following sentences from \{Source Language\} to \{Target Language\}. \{Source Text\}}
    \item \textbf{Prompting}: \textit{\{Source   Language\}: \{Source Text\} \textbackslash{}n\{Target Language\} Translation:}
\end{itemize}
\paragraph{Cross-lingual Summary}
\begin{itemize}
    \item \textit{Please write a concise summary of the text, return your responses with one line that cover the key points of the text.\textbackslash{}n Input: \{Source Text\} \textbackslash{}n Output: \{Target Language\} Summary:}
\end{itemize}
\paragraph{POS tagging}
\begin{itemize}
    \item \textit{Please provide the POS tags for each word in the input sentence. The input will be a list of words from the sentence. The output format should be a JSON array, where each key-value pair includes a word from the input list and its corresponding POS tag from the set of labels: {[}\{PoS Tags\}{]}\textbackslash{}nNote: Your response should include only a JSON array, presented in the order that the words appear in the input   sentence. Each key-value pair should contain the word and its POS label.}
\end{itemize}

\subsection{Success Rate of Vanilla Prompting} \label{sec:success}
We randomly selected 500 sentences from the Flores-101 benchmark for four translation directions and randomly used ten diverse prompts for the corresponding translations. In addition to prompts proposed by \citet{bawden-yvon-2023-investigating}, the used prompts are as follows:
\begin{itemize}
    \item \textit{Translate the following sentences from \{Source Language\} to \{Target Language\}.}
    \item \textit{Can you help me translate this sentence from \{Source Language\} to \{Target Language\}?}
    \item \textit{What is the translation of sentence in \{Target Language\}? \{Source Text\}}
    \item \textit{Infill in \{Target Language\}: \{Source Text\}}
    \item \textit{Please provide machine translation text for me.\textbackslash{}n\{Source Language\}: \{Source Text\}\textbackslash{}n \{Target Language\}:}
\end{itemize}
To determine whether a translation was generated (regardless of its translation quality), we employed automated tools\footnote{\url{https://github.com/Mimino666/langdetect}} and then checked the results by human. We then calculated the success rate of translation requests. The results, which will be detailed in table below, indicate that foundation LLMs exhibit considerable instability in generating the target language when presented with different prompts. In most cases, the LLMs failed to produce the desired language output.

\begin{table}[H]
\scalebox{0.9}{
\begin{tabular}{lcccc}
\toprule
                       & \multicolumn{1}{l}{\textbf{En-Zh}} & \multicolumn{1}{l}{\textbf{En-Uk}} & \multicolumn{1}{l}{\textbf{Zh-Pt}}        & \multicolumn{1}{l}{\textbf{De-Fr}}    \\               
\midrule
Success Rate $\uparrow$    & 28.6\% &	21.6\%	& 39.6\%	& 37.4\%                                \\
\bottomrule
\end{tabular}
}
\caption{Success Rate of translation prompting with different instructions.}
\end{table}

\subsection{Pseudo Labels}\label{sec:dic}
We collected lexical data from various sources, including Wiktionary\footnote{\url{https://www.wiktionary.org/}}, MUSE\footnote{\url{https://github.com/facebookresearch/MUSE\#ground-truth-bilingual-dictionaries}}\cite{DBLP:conf/iclr/LampleCRDJ18}, and PanLex\footnote{\url{https://panlex.org/source-list/}}, and merge their entries into one dictionary. \cref{tab:lex} presents the detailed information of compiled dictionaries.
\begin{table}[H]
\centering
\scalebox{0.9}{
\begin{tabular}{crc}
\toprule
\textbf{Languages} & \multicolumn{1}{l}{\textbf{\# Entries}} & \textbf{Source}  \\
\midrule
En-Zh              & 31,443                                  & Wiktionary, MUSE \\
Zh-En              & 13,667                                  & MUSE\\
En-Uk              & 32,685                                  & MUSE\\
Uk-En              & 34,888                                  & MUSE\\
De-Fr              & 61,527                                  & Wiktionary, MUSE \\
Pt-Zh              & 322,987                                 & PanLex          \\
\bottomrule
\end{tabular}
}
\caption{The detailed statistical information of the dictionary used for constructing Pseudo Priors.}
\label{tab:lex}
\end{table}

\end{document}